\documentclass[10pt]{cai}

\graphicspath{{figs/}}

\usepackage{lipsum}
\usepackage{abstract}
\usepackage{times}
\usepackage{latexsym}
\usepackage{enumitem}
\usepackage{wrapfig}

\usepackage[T1]{fontenc}

\usepackage[utf8]{inputenc}

\usepackage{microtype}

\usepackage{inconsolata}

\usepackage{graphicx}
\usepackage{caption}
\usepackage{subcaption}
\usepackage{amsmath}

\begin{document}
\linepenalty=100
\volumeheader{36}{0}
\begin{center}

  \title{Parameter-Efficient Methods for Metastases Detection from Clinical Notes}
  \maketitle

  \thispagestyle{empty}

    

  \begin{tabular}{cc}
    Maede Ashofteh Barabadi\upstairs{\affilone,*}, Xiaodan Zhu\upstairs{\affilone}, Wai Yip Chan\upstairs{\affilone}, Amber L. Simpson\upstairs{\affiltwo}, Richard K.G. Do \upstairs{\affilthree}
   \\[0.25ex]
   \small \upstairs{\affilone} Department of Electrical and Computer Engineering and  Ingenuity Labs Research Institutes \\Queen's University, Kingston, ON, Canada \\
   \small \upstairs{\affiltwo} School of Computing and Department of Biomedical and Molecular Sciences\\Queen's University, Kingston, ON, Canada \\
   \small \upstairs{\affilthree} Department of Radiology, Memorial Sloan Kettering Cancer Center, New York, NY, USA \\
  \end{tabular}
  
  \emails{
    \upstairs{*}maede.ashofteh@gmail.com 
    }
  \vspace*{0.2in}
\end{center}

\begin{abstract}
Understanding the progression of cancer is crucial for defining treatments for patients. The objective of this study is to automate the detection of metastatic liver disease from free-style computed tomography (CT) radiology reports. Our research demonstrates that transferring knowledge using three approaches can improve model performance. First, we utilize generic language models (LMs), pre-trained in a self-supervised manner. Second, we use a semi-supervised approach to train our model by automatically annotating a large unlabeled dataset; this approach substantially enhances the model's performance. Finally, we transfer knowledge from related tasks by designing a multi-task transfer learning methodology. We leverage the recent advancement of parameter-efficient LM adaptation strategies to improve performance and training efficiency. Our dataset consists of CT reports collected at Memorial Sloan Kettering Cancer Center (MSKCC) over the course of 12 years. 2,641 reports were manually annotated by domain experts; among them, 841 reports have been annotated for the presence of liver metastases. Our best model achieved an F1-score of 73.8\%, a precision of 84\%, and a recall of 65.8\%.
\end{abstract}

\begin{keywords}{Keywords:}
Parameter-Efficient Tuning, Pre-trained Language Models, Metastases Detection.
\vspace{-4mm}
\end{keywords}
\copyrightnotice

\section{Introduction}
\label{intro}
\vspace{-2mm}
Progression of metastatic disease is often the primary cause of cancer-related death~\cite{menezes2016detecting}, thus early detection of metastasis is important for selecting targeted and other therapies. In the liver, for example, metastases can be treated more effectively when discovered early. 
Understanding the spatial and temporal patterns of metastases distribution would help radiologists more accurately interpret CT images for the existence of any metastasis. In order to extract the patterns, a comprehensive analysis of large-scale clinical data is necessary, but this is difficult given the unstructured nature of most electronic health records.
Since cancer patients receive many CT scans as part of care, the corresponding reports contain rich data that can be mined for cancer recurrence and progression.
Annotating CT reports requires domain expertise and is costly and time-consuming to perform manually on a large scale. Therefore, automation of metastatic site detection from radiology reports can substantially advance studying and treating cancer progression.

Since the amount of human-annotated data is limited, training large models has a high risk of overfitting. However, the strategy of pre-training large LMs followed by task-specific fine-tuning allows us to tailor to a new task using a small task-specific dataset. While full fine-tuning is the conventional adaptation paradigm, parameter-efficient tuning has recently been shown to achieve comparable performance by adapting only a small percentage of the parameters~\cite{prompt-tuning-v2}. However, they have not received enough study in medical applications. In this work, we adapt a pre-trained LM through fine-tuning and prompt-tuning --- a typical parameter-efficient tuning approach --- to the task of detecting liver metastases. We also employ a semi-supervised approach by leveraging a dataset annotated by another machine learning model.

The data used in this study were collected at Memorial Sloan Kettering Cancer Center (MSKCC) from July 2009 to May 2022 by waiver of informed consent and follows a structured departmental template, which includes a separate header under the findings section for each organ and an impression section that summarizes key observations. Previous studies have shown promising results by exploiting all sections related to the organ of interest~\cite{do2021patterns, batch2022developing}, but their applicability is limited to radiology reports with a similar structure. To reduce the reliance on the report format and increase the applicability of the proposed methods to a wider variety of radiology reports, only the impression section is used as input.

Our main contributions are as follows: 
(1) We propose to use parameter-efficient tuning --- the soft prompt tuning --- to solve the problem and demonstrate that it outperforms full fine-tuning when only a small manually curated dataset is available.
(2) Our introduced methods only require the presence of an impression section (i.e., free text), which is a common practice in radiology reports, so their applicability can be extended to most radiology reports.
(3) We train BERT on a large-scale, automatic-annotated dataset, which leads to higher performance than training on a small, human-annotated dataset.
(4) We also present a multi-task transfer learning method based on prompt-tuning which improves performance moderately.

\section{Dataset and Problem Description}
\vspace{-2mm}
\paragraph{\textbf{Dataset.}} The data used in our experiments were gathered at MSKCC from July 2009 to May 2022. The entire collected data was split into two specific datasets. The first dataset was annotated by five radiologists, for the presence of liver metastases. They were instructed to read all reports available for each patient, including future reports, before deciding on the presence or absence of metastases at the time of each report. Further details of the annotation process can be found in ~\cite{batch2022developing}. This process resulted in 2,641 annotated reports from 314 patients. Data were partitioned into training (20\%), validation (30\%), and testing samples (50\%) by patients. Half of the dataset records are allocated for testing, aiming to ensure evaluation quality. The remaining 50\% for training and validation reflects the scarcity of data in real-life applications. 

The second dataset records are automatically annotated with a fine-tuned BERT model trained following the method in~\cite{do2021patterns}. The annotating model had access to the dedicated organ section and impression section. This automatic-annotated dataset consists of 1,032,709 radiology reports from 192,650 patients and has annotations for 13 organs: \textit{liver}, \textit{lungs}, \textit{pleura}, \textit{thoracic nodes}, \textit{spleen}, \textit{adrenal glands}, \textit{renal}, \textit{abdominopelvic nodes}, \textit{pelvic organs}, \textit{bowel/peritoneum}, and \textit{bones/soft tissues}. Since automatic-annotated labels are noisy,
the evaluation of all trained models was done on the human-annotated test set, regardless of their training data.

\begin{wraptable}{R}{8cm}
       \caption{Examples of Impression Text}
    \begin{tabular}{lp{7cm}}
    \hline
         1 & Since <date>, 1. Stable collection the hepatic resection margin. \\
         \midrule
         2 & Since <date>, no interval changes. \\
         \midrule
         3 & Since <date>, 1. Status post right hemicolectomy with mural soft tissue thickening or retained material in the colon just distal to the anastomosis. Correlation with endoscopy recommended. Email sent to <person>. 2. Status post partial hepatic resection with no evidence of new hepatic lesion. Reduced size of fluid adjacent to resection margin consistent with reduced postoperative change. 3. Stable tiny pulmonary nodules.\\
    \bottomrule
    \end{tabular}
    \label{sample_impressions}
\end{wraptable}

\paragraph{\textbf{Problem Formulation.}} We formulate the problem of detecting liver metastasis from the impression section of a radiology report as a binary classification task. Our model input is the impression section of the report to closely mimic the real-life setup.
Table \ref{sample_impressions} shows some sample impression texts. Some of the texts are relatively non-informative, like example 2, while others are more detailed. We denote the training set as $\{(x, y)\}$, where $x$ is an impression text, and $y \in \{0, 1\}$ is the ground-truth label when $1$ indicates the presence of liver metastasis and a $0$ indicates no liver metastasis. We use $p_{\theta}(x)$ to denote the probability of a positive class predicted by a model parameterized by $\theta$.

\section{Related Work}
\label{related_work}
\vspace{-2mm}
\paragraph{\textbf{Analyses of Cancer Patient Clinical Records.}}
Previous research on detecting metastasis has analyzed CT images~\cite{vorontsov2019deep}. However, using CT reports instead of images provides more comprehensive information, as radiologists consider a patient's medical history when interpreting the images. Researchers have applied a wide range of natural language processing (NLP) techniques to interpret CT reports, from rule-based methods~\cite{10.1200/CCI.21.00030} to classical machine learning algorithms~\cite{causa2022natural, chen2018integrating} to deep neural networks~\cite{kehl2019assessment}. For example, the authors in~\citep{do2021patterns} used both classical NLP methods --- a TF-IDF feature extractor and SVM/random forest classifiers --- and BERT to detect metastatic sites from structured radiology reports. Another study utilized long short-term memory (LSTM) and convolutional neural network (CNN) and found that accessing previous reports is beneficial in detecting metastasis~\cite{batch2022developing}. Although these two works show promising results, their data follow the previously mentioned departmental template, so the application of their models is limited to reports from a very specific institute. To the best of our knowledge, our work is the first to address this limitation by performing metastasis detection based solely on the impression section.

\paragraph{\textbf{Parameter-Efficient Tuning for Classification.}}
The most common paradigm of adapting general pre-trained LMs to a specific task is fine-tuning, in which all parameters of the pre-trained model are updated using data for the downstream task. However, as LMs have grown inexorably larger, the cost of fine-tuning has become burdensome. To address this issue, researchers have introduced parameter-efficient methods that freeze (do not update) all or part of the LM parameters. These methods either fine-tune only a small portion of model parameters, such as BitFit~\cite{bitfit} and child-tuning~\cite{child-tuning}, or introduce new parameters and train them from scratch, such as adapter-tuning~\cite{adapter-tuning}. 
Prompt-tuning is a parameter-efficient method that prepends extra tokens to the keys and values in the attention layers~\cite{prompt-tuning}. The concept of prompt-tuning was first introduced in~\cite{prefix-tuning}, which demonstrated promising results on natural language generation tasks. Subsequently, \cite{prompt-tuning} employed the method (with some modifications) on classification tasks by translating them into a text-to-text format. It yielded comparable performance to fine-tuning when the model size exceeded one billion parameters. P-tuning v2~\cite{prompt-tuning-v2} further extended this research to natural language understanding (NLU) by adding a trainable layer on top of the LM. Their proposed architecture performs comparably with fine-tuning over different scales. In this work, we use P-tuning v2 to train a classifier for metastasis detection.

\paragraph{\textbf{Parameter-Efficient Multi-Task Transfer Learning.}}
Multi-task transfer learning is a strategy that enhances the performance of models on a target task by transferring useful knowledge from related tasks. Prior studies have investigated multi-task approaches that are compatible with prompt-tuning. For example, SPoT~\cite{vu2021spot} suggests initializing the downstream task prompts with prompts that have been tuned on a mixture of related tasks. Meanwhile, HyperPELT~\cite{zhang2022hyperpelt} trains a hypernetwork that generates trainable parameters for the main model, including prompt tokens. Another approach, ATTEMPT~\cite{asai2022attempt}, learns prompts for all the source tasks and then creates an instance-wise prompt for the target task by combining the source tasks' prompts and a newly initialized prompt using an attention block. We will discuss how our method is different from theirs in the methodology section.

\section{Methodology}
\label{methodology}
\vspace{-2mm}

\begin{figure}[t]
\centerline{\includegraphics[scale=0.15]{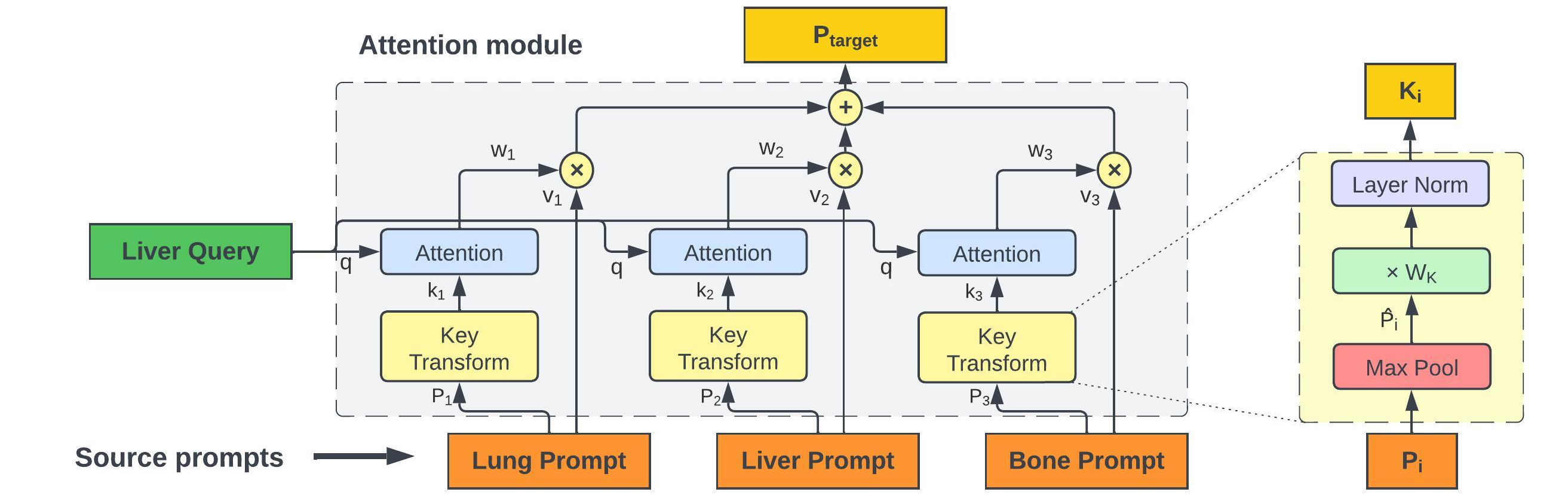}}
\caption{Proposed multi-task soft prompt architecture.}
\label{fig:multi-task}
\vspace{-6mm}
\end{figure}

To address the scarcity of manually annotated data, we employ several strategies. Firstly, we utilize pre-trained LMs by adapting prompt-tuning to reduce the risk of overfitting. Secondly, we augment the training data by automatically annotating a large dataset that would be challenging to label manually. Lastly, we present a multi-task transfer learning framework that allows the model to leverage information from other organs. This method builds upon the prompt-tuning approach and formulates the final target task prompt as a linear combination of source prompts. Figure \ref{fig:multi-task} illustrates this process. We have 13 source prompts, $P_1, P_2, ..., P_{13}$, but only three of them are shown in Figure \ref{fig:multi-task} for the sake of demonstration. The source prompts were learned using P-tuning v2~\cite{prompt-tuning-v2} on the source tasks of detecting metastasis in different organs, including the liver. P-tuning v2 and our prompt attention mechanism are described in detail in the following sections.

\paragraph{\textbf{Prompt-Tuning.}}
Assume we have an encoder building on any Transformer-based LM with a classifier layer on top of the last representation layer. We denote this architecture as $p_{\theta,\theta_c} (x)$, where $\theta$ and $\theta_c$ refer to the LM parameters and classification head parameters, respectively. In fine-tuning, we tune all parameters by optimizing $min_{(\theta, \theta_c)} \mathrm{BCELoss}(p_{\theta,\theta_c} (x), y)$ over all ($x$, $y$) pairs from the training set. $\mathrm{BCELoss}$ refers to binary cross-entropy loss, the conventional loss function for classification problems. In P-tuning v2~\cite{prompt-tuning-v2}, prompt tokens are prepended to the keys and values of the attention modules in all transformer layers, as described in Equation \ref{eq:p-tuning-v2}. $h_i$ is the output of $i$-th transformer encoder layer, and $f_i$ is the output of the attention layer in the same transformer block, while $q_i$, $k_i$, and $v_i$ denote the query matrix, key matrix, and value matrix in the $i$-th layer, which are obtained by transferring the last layer output with  $W_i^Q$, $W_i^K$, and $W_i^V$ matrices to new latent spaces. Before computing attention, we concatenate key prompt tokens $p^K_i\in\mathbf{R}^{d_m\times pl}$ and value prompt tokens $p^V_i\in\mathbf{R}^{d_m\times pl}$ with the key and value matrices where $pl$ refers to prompt length.

\vspace{-1mm}
\begin{equation}
\label{eq:p-tuning-v2}
\begin{gathered}
q_i, k_i, v_i = W_i^Q h_{i-1}, W_i^K h_{i-1}, W_i^V h_{i-1}\\
f_i = \mathrm{MultiHeadAttention}(q_i, [p^K_i; k_i], [p^V_i; v_i])\\
\end{gathered}
\end{equation}

The LM parameters are frozen during prompt-tuning. The only trainable parameters are the prompt tokens and the classification head. So, we can formulate the prompt-tuning optimization problem as $min_{(\theta_c, p^K, p^V)}\mathrm{BCELoss}(p_{\theta, \theta_c, p^K, p^V} (x), y)$. Depending on the prompt length, P-tuning v2 reduces the trainable parameters to ~0.5-2\% of that of full fine-tuning. We did not observe any improvement from reparameterization and thus we learned prompt tokens directly.

\paragraph{\textbf{Attentional Mixture of Prompts.}}
After obtaining source prompts from the prompt-tuning method, we interpolated them to form a new prompt for the target task using an attention module (Figure \ref{fig:multi-task}). The source prompt weights $w_i$ were determined by the attention between the target task query $q$ and keys $k_i$. To generate keys, we first reduce the dimensionality of the source prompts by max pooling and make a compact representation $\hat{P_i} \in \mathbf{R}^{d_m}$, where $d_m$ represents the LM hidden size, which is 768 for BERT-base. We then map the max-pooled source prompts to a new space via transformation matrix $W_K$, and apply layer normalization to prevent gradients from becoming excessively large. The attention module calculates the target prompt using Equation \ref{eq:attention_formula}, where $e$ and $n$ are Euler's number and number of source tasks, respectively.

\vspace{-1mm}
\begin{equation}
\begin{split}
\label{eq:attention_formula}
k_i = \mathrm{LayerNorm}(W_K \hat{P_i})\quad\quad w_i = \dfrac{(q_\cdot k_i/(e\cdot d_{m}))^2}{\Sigma_{j=1}^{n} (q\cdot k_j/(e\cdot d_{m}))^2}\quad\quad P_{target} = \Sigma_{j=1}^{n} w_j P_j
\end{split}
\end{equation}

The conventional attention method uses \textit{softmax} to normalize weights, which tends to assign a high weight to the liver source prompt and very small weights to other source prompts. This impedes the effective transfer of knowledge between tasks. Instead, we apply a degree-2 polynomial kernel to produce more evenly distributed weights. We scale the dot product of the key and query to make the result independent of the model's hidden size. $W_K$ and $q$ are trainable parameters of the attention block, while other components, including source prompts, remain frozen. We prepend $P_{target}$ tokens to all model layers and pass input through LM to compute the model's output.

In the multiple target tasks case, the attention module parameters can be shared. After training is finished, $P_{target}$ can be calculated once and saved. Our method is different from ATTEMPT~\cite{asai2022attempt}, which requires both the attention module and source prompts during inference in order to compute its instance-dependent attention query, leading to more computation and storage. Our method operates like P-tuning v2 during inference with no additional parameters or computation steps.

\vspace{-1mm}

\section{Experiments and Results}
\vspace{-2mm}
\label{experiments}
\paragraph{\textbf{Experiment Setup.}}
We evaluated all models on the human-annotated test set. We fine-tuned BERT using both the human-annotated and automatic-annotated datasets. Additionally, we obtained prompt-tuned models on both datasets, which also leveraged BERT-base as the backbone LM. Our Multi-task model was solely trained on the automatic-annotated data, as it provided metastasis annotation for multiple organs.
The implementation of P-tuning v2 was based on the source code provided by the authors\footnote{https://github.com/THUDM/P-tuning-v2}. The models were trained for a maximum of 1,000 epochs on human-annotated data and 10 epochs on automatic-annotated data. The best checkpoint was selected based on the F1-score on the validation set.
To address the problem of data imbalance, we upsampled the positive class to balance the number of samples per class. We found the best batch size, learning rate, and prompt length, when applicable, based on F1-score on the development set.

\label{results}

\begin{table}
\renewcommand\arraystretch{0.9}
    \caption{Performance of different models. The Val. F1 and Test F1 refer to F1-scores on the validation and test set, respectively, while \textit{manual} and \textit{automatic} refer to using manually annotated and automatically annotated training data, respectively. The improvement of the multi-task model over both the fine-tuning and prompt-tuning is statistically significant (p < 0.01) under the one-tailed paired T-test.}

    \label{tab:result}
   
    \centering
    \setlength{\tabcolsep}{2pt}
\setlength{\belowcaptionskip}{-2mm}
    \resizebox{0.9\linewidth}{!}{

    \begin{tabular}{p{2.5cm}| c || c|ccc | r}

    \toprule

        Method & Training data & Val. F1 & Test F1 & Precision & Recall & \# Tunable param\\
        \hline

        Fine-tuning & manual & 75.8 & 69.0 & 74.3 & 64.3 & 109M\\
        Prompt-tuning & manual & 75.6 & 71.9 & 69.1 & 74.9 & 1,236K\\             

      \hline
        Fine-tuning & automatic & 79.7 & 73.4 & 89.7 & 62.1 & 109M\\
        Prompt-tuning & automatic & 79.6 & 73.3 & 86.0 & 63.8 & 1,624K\\
        Multi-task model & automatic & 79.7 & \textbf{73.8} & 84.0 & 65.8 & 2,218K\\
      \bottomrule

    \end{tabular}

    }
\end{table}

\paragraph{\textbf{Experiment Results.}}
The performance of the models is summarized in Table \ref{tab:result}. 
On manually annotated data (\textit{manual}), prompt-tuning improves the test F1-score by almost three points (from 69.0\% to 71.9\%) with only 1\% tunable parameters compared to the fine-tuning (1.2M vs. 109M), showing that prompt-tuning performs better in the low-data setting, where only a limited amount of (manually annotated) training data is available. This can be attributed to the fact that prompt-tuning has far fewer parameters, making it less prone to over-fitting, which can be seen from the difference in performance between the validation and test set. 

When the amount of training data is much larger using automatically annotated data (\textit{automatic}), with around 1 million samples, fine-tuning and prompt-tuning perform similarly. In this case, prompt-tuning is still preferable, since it is computationally more efficient during training and can be served in shared mode with other tasks with considerably reduced memory (1.6M tunable parameters vs. 109M in fine-tuning). This benefit will be more significant as the pre-trained models continue to grow significantly larger every year.

Our proposed multi-task approach surpasses both prompt-tuning and fine-tuning. These outcomes suggest that transferring knowledge from related tasks in the medical domain can enhance the performance of the prompt-tuning method while maintaining parameter efficiency. Our experiments only utilized 13 source tasks, and incorporating more related tasks may result in greater improvements.

Our observation from Table \ref{tab:result} reveals that the models trained on automatically-annotated data outperform those on human-annotated data for both fine-tuning and prompt-tuning methods. This suggests that even if we use parameter-efficient methods, a few hundred annotated records are not sufficient to obtain high performance for liver metastasis detection from impression text. While manually annotating large datasets is a time-consuming and resource-intensive approach, automatically annotating data using a model that has access to more information from the input report is a low-cost alternative that we proved worthy of pursuit.

\section{Conclusion}
\vspace{-2mm}
\label{conclusion}
In this paper, we propose metastatic liver identification from free-style radiology reports by removing restrictive assumptions about the report structure.
Our results indicate that soft prompt-tuning, as a typical parameter-efficient method, surpasses fine-tuning in the low-data setting and achieves comparable results with a large train set. It implies that prompt-tuning can be used to build more efficient models without sacrificing performance. Additionally, we proposed a multi-task transfer learning framework and found it to improve the performance of metastasis detection by leveraging information from related tasks. We also demonstrated the usefulness of training on large automatically annotated data via a semi-supervised approach. This suggests that artificially annotating large datasets is an effective solution to overcome the challenge of limited labeled data in tasks with similar settings. These techniques have the potential to be applied to other tasks in the medical domain that have a similar setup.

\section*{Acknowledgements}
\vspace{-2mm}
This research is supported by the Vector Scholarship in Artificial Intelligence, provided through the Vector Institute.~\footnote{https://vectorinstitute.ai/} The research is partially supported by NSERC Discovery Grants.
\vspace{-2mm}
\appendix

\printbibliography[heading=subbibintoc]

\end{document}